\RequirePackage{snapshot}
\documentclass[10pt,twocolumn,letterpaper]{article}

\usepackage{iccv}
\usepackage{booktabs}
\usepackage{times}
\usepackage{epsfig}
\usepackage{graphicx}
\usepackage{amsmath}
\usepackage{amssymb}
\usepackage{afterpage} 
\usepackage[font=small]{caption}
\usepackage{cite}
\usepackage{verbatim}

\graphicspath{ {images/} }

\usepackage[pagebackref=true,breaklinks=true,letterpaper=true,colorlinks,bookmarks=false]{hyperref}

\iccvfinalcopy 

\def\iccvPaperID{0001}

\ificcvfinal\fi
\begin{document}

\title{
Are we asking the right questions in MovieQA?
}

\author{
Bhavan Jasani$^{1}$ \qquad Rohit Girdhar$^{1}$ \qquad Deva Ramanan$^{1,2}$ \\
$^{1}$The Robotics Institute, Carnegie Mellon University \quad $^{2}$Argo AI \\
{\small \url{http://BhavanJ.github.io/MovieQAWithoutMovies}}
}

\maketitle

\ificcvfinal\thispagestyle{empty}\fi

\begin{abstract} Joint vision and language tasks like visual question answering are fascinating because they explore high-level understanding, but at the same time, can be more prone to language biases. In this paper, we explore the biases in the MovieQA dataset and propose a strikingly simple model which can exploit them. We find that using the right word embedding is of utmost importance. By using an appropriately-trained word embedding, about half the Question-Answers (QAs) can be answered by looking at the questions and answers alone, completely ignoring narrative context from video clips, subtitles, and movie scripts. Compared to the best published papers on the leaderboard, our simple question+answer only model improves accuracy by 5\% for video + subtitle category, 5\% for subtitle, 15\% for DVS and 6\% higher for scripts.
\end{abstract}

\section{Introduction}

Language has long been an integral part of visual understanding. From objects~\cite{imagenet_cvpr09,DBLP:journals/corr/LinMBHPRDZ14} to human actions~\cite{DBLP:journals/corr/KayCSZHVVGBNSZ17}, categorization of visual data has lead to rapid developments in computer
vision. However, language is particularly transformative because it can be applied to domains beyond simple classification,
such as image captioning~\cite{msr-vtt-a-large-video-description-dataset-for-bridging-video-and-language} and Visual Question-answering (VQA)~\cite{DBLP:journals/corr/AntolALMBZP15}. Indeed, VQA has arguably emerged as a now-standard vision task, primarily due to its flexibility and standardized evaluation.

{\noindent \bf MovieQA:} QA tasks are particularly intriguing for videos, where they can explore cognitive storytelling concepts (such as intentions and goals) difficult to extract from static images.
Unsurprisingly, there have been considerable efforts in bridging the gap between language and spatio-temporal understanding of videos~\cite{lei2018tvqa,MovieQA}.
To that end, a recently released dataset, MovieQA~\cite{MovieQA}, extends the VQA philosophy to videos,
by collecting short real-world movie clips, along with subtitles and wiki-plots, and defining multiple choice questions on them.
It has 5 categories for the QA task based on the information used:
1) movie clips + subtitles 2) movie subtitles 3) movie scripts 4) DVS (descriptive video services) 5) Wikipedia movie plots (wiki-plots). The first category is based on the combination of visual and text data, whereas the remaining 4 are purely text-based tasks. 
While there has been a significant amount of work in this direction, most methods \cite{Na_2017_ICCV,Wang2018,liua2a,ECCV_2018_Dual_Attention_Memory} do not make strong use of visual features and instead rely heavily on language-based cues such as subtitles or wiki-plots.
This raises the question: are our video models unequipped to truly understand videos, or is the MovieQA task unfairly biased against actually needing visual information?

{\noindent \bf WikiWord embeddings:} In this work, we explore this question in detail. We propose a strikingly simple approach that extracts average-pooled word embeddings of the question and each answer and reports the answer with the best correlation. We train our word embedding model -- named {\em WikiWord} embeddings -- on unsupervised Wikipedia plots, to capture the narrative structure of movie plots. We find that this simple model outperforms {\em all} reported methods on MovieQA~\cite{MovieQA} test set. This includes models that use subtitles, scripts, and videos, while our naive model uses {\em only} the question and answer. We have submitted our results to the test evaluation server, and are ranked first in four out of five categories at the time of submission of this paper.

{\noindent \bf The role of plots:} It is worth noting the one category that we do {\em not} win is plot-synopsis (wiki-plots), where the current state-of-the-art is quite high (85\%). This is explained by the fact that the question and answers were {\em constructed} by inspection of movie plots from Wikipedia. This category provides aligned training examples of $\{$(question,answer,plot$)_i\}$ tuples for supervised learning, which can be exploited by powerful language models that are trained on such aligned data~\cite{blohm2018comparing}. In contrast, we learn embeddings in an unsupervised fashion from {\em un}aligned movie plots $\{$plot$_i\}$.
This information is freely available in all the 5 benchmark category protocols. Our results demonstrate that {\em unsupervised} learning of word-embeddings from {\em un}aligned movie plots still captures a rich amount of narrative structure about the movies of interest.

{\noindent \bf Source of bias:} The source of language bias might be explained by the procedure used to generate the benchmark QAs: Amazon Turkers generate candidate QAs by reading the movie plots {\em without} watching the movies. Movie clips are later programmatically aligned to movie plot lines and the questions. Moreover, we find that for many QAs, words and characters from the relevant movie plots are included in the correct answer, but not included in the incorrect answers.
This may make it easier to pick out the correct answer by simply looking at the question and answers.

{\bf \noindent Why is this relevant for vision?} Because our central technical contribution is a novel language baseline model, one might argue that it is not relevant for a vision audience. We believe it is crucial to ensure that strong baselines are introduced for the tasks at hand, to ensure meaningful progress is made. Hence, we feel that our results are very relevant for the MovieQA community and future joint language-vision datasets. Additionally, our method naturally generates a partition of the data that is free of such trivial biases and can potentially be used for further progress in video-language modeling.

\begin{figure} \begin{center}
\includegraphics[width=\linewidth,keepaspectratio]{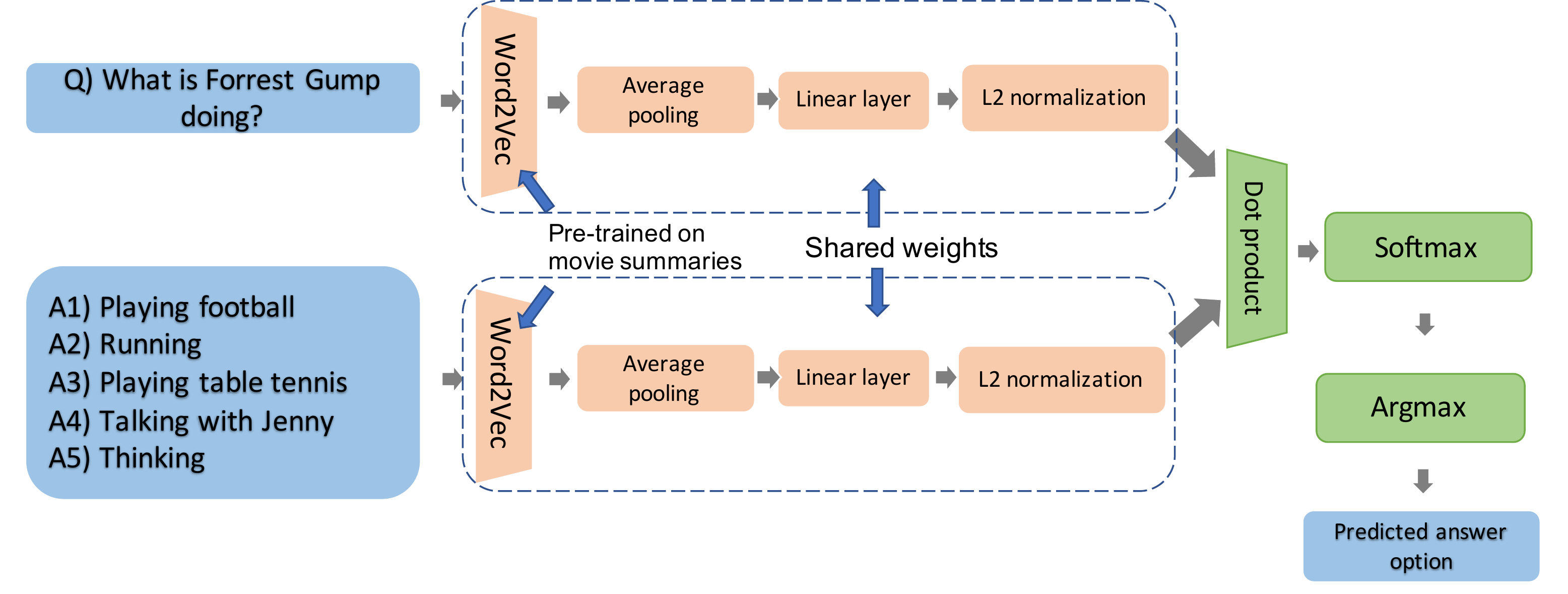}

    \end{center}
    \caption{{\bf WikiWord Embedding model.} It takes as input the question and 5 answer choices. For every word in the sentence (of question and answer choices), a 300D word embedding is computed using word2vec. This word2vec is pretrained on movie plots and its weights are kept fixed. The word embedding is average pooled to get a sentence level vector and then passed through a linear layer (initialized as an identity matrix) to get another 300D vector which is then L2 normalized. Dot product similarity is computed for the 300D representation of question and the 5 answer choices, and the one with the highest value is picked as the model's predicted answer option.
} 
\label{fig:our_model}
\end{figure}

\section{Related Work}
{\noindent \bf Video and language:} Joint learning of language and vision 
has been explored in various ways.
This includes movie descriptions\cite{lsmdc}, video understanding through fill in the blank\cite{maharaj2017dataset}, video retrieval \cite{lsmdc2016MovieAnnotationRetrieval}, character co-referencing \cite{DBLP:journals/corr/RohrbachRTOS17} and image captioning\cite{msr-vtt-a-large-video-description-dataset-for-bridging-video-and-language}. Most previous works have focused on using movies~\cite{lsmdc2016MovieAnnotationRetrieval, DBLP:journals/corr/RohrbachRTOS17, 2018arXiv180605341H}, because they provide time synchronized audio, subtitles and videos.

{\noindent \bf Visual QA task:}
Question answering
provides an easy and unambiguous evaluation metric for joint language and vision tasks. The task is to predict the correct answer from a list of options for a given question based on a story, which provides the context.
Many visual question answering datasets have been recently released, including image-based question answering datasets like VQA~\cite{DBLP:journals/corr/AntolALMBZP15}, and more recently, video-based QA.
This includes datasets like MovieQA~\cite{MovieQA}, constructed from movies, TVQA~\cite{lei2018tvqa}, constructed from TV series and TGIF QA~\cite{DBLP:journals/corr/JangSYKK17}, constructed from GIFs. Additionally, there has been work on reading comprehension~\cite{DBLP:journals/corr/HermannKGEKSB15}, which are the purely language-based QA datasets.

\section{Our Approach: WikiWord Embeddings}

{\noindent \bf Classic formulations:}  Typical QA task can be formalized as triplets consisting of the reference passage (to be comprehended), a question, and the possible answers (5 choices in case of MovieQA). Contemporary QA systems create a scoring function that iterates over all putative answers, conditioned on the question and reference passage, returning the highest-scoring answer.

{\bf \noindent Default word2vec:} Let us first review the basic Visual QA framework provided in the MovieQA benchmark~\cite{MovieQA}, which forms the basis for our proposed solution. Of particular relevance is the default word2vec, which is trained on 1400 Wikipedia movie plots, including movies in the train split, test split, and movies outside MovieQA. It is important to note that the word embeddings are learnt from movie plots in an unsupervised way, without looking at the corresponding questions and ground-truth answers. 

{\noindent \bf WikiWord embedding model:} Our crucial modification trains a word2vec embedding {\em only} on movies present in MovieQA (train and test splits), a strict subset of the data used to train the default word2vec embedding. We call our embedding WikiWords. We use it in a simple pipeline (Fig.~\ref{fig:our_model}) that makes use of {\em only} questions and answers, ignoring any reference passage, subtitles, or videos. Specifically, we compute a sentence-level embedding for each question and answer by average pooling WikiWord embeddings. We then select the answer with the highest (weighted) similarlity to the question. Note that the linear reweighting is the only component of our model that is trained on question-answer pairs. We also provide experimental results for a variant of our model without any linear tuning, which is trained without {\em any} question-answer supervision.

\section{Experiments}

\begin{table*} \centering
\small
\begin{tabular}{ccc}
\begin{tabular}{|l c|} 
\toprule
Leader board submission & Subtitles \\
\midrule
 Our QA-only model       &  {\bf 44.01}    \\
Speaker Naming in Movies\cite{Speaker_naming}  & 39.36 \\
\bottomrule
\end{tabular}
&
\begin{tabular}{|lc|} 
    \toprule
    Leader board submission & DVS  \\
    \midrule
     Our QA-only model      &  {\bf 49.65}    \\
MovieQA benchmark\cite{MovieQA}  & 35.09 \\
   \bottomrule
\end{tabular}
&
\begin{tabular}{|lc|} 
\toprule
Leader board submission & Scripts  \\
\midrule
 Our QA-only model       &  {\bf 45.49}    \\
Read Write Mem. Net.\cite{Na_2017_ICCV}  & 39.36 \\
\bottomrule
\end{tabular}
\end{tabular}
\caption{MovieQA leaderboard for Subtitles, DVS, and Scripts categories at the time of submission along with the second best submissions.
}
\label{table:leaderboard}
\end{table*}

{\noindent \bf Leaderboard results:} The dataset is divided into train, validation (val) and test splits. 
We report ablation experiments on the val set. The test results are obtained from the official server\footnote{\url{http://movieqa.cs.toronto.edu/leaderboard/}}, and are reported in Tables~\ref{table:leaderboard} and \ref{table:leaderboard_video} for the various categories.
Table \ref{table:LMN_results} shows the performance of different input modalities (QA only, subtitles, videos and videos+subtitles) for the top model on the leaderboard with publicly released code~\cite{Wang2018}. 
Our results dominate past work by a significant margin (5 percent), while using strictly less information for learning word embeddings and ignoring reference material such as subtitles, DVS, or scripts.

\begin{table} \centering
\small
\setlength{\tabcolsep}{3pt}
\begin{tabular}{ lc } 
\toprule
Leader board submission & Movie: Video+Subtitles  \\
\midrule
Our QA only model      &   {\bf 46.98}    \\
New method to optimize all   &  \\
 MEM network (anonymous) & 45.31 \\
Multimodal dual attention memory\cite{ECCV_2018_Dual_Attention_Memory}  &  41.41    \\
\bottomrule
\end{tabular}\caption{MovieQA leaderboard for Video+Subtitles category at the time of submission along with previous best anonymous and published results.
}
\label{table:leaderboard_video}
\end{table}

\begin{table} \centering
\small
\setlength{\tabcolsep}{3pt}
\begin{tabular}{ lcccc } 
\toprule
Modality           & Google~\cite{DBLP:journals/corr/MikolovSCCD13} & MovieQA~\cite{MovieQA} & Our best w2v \\
\midrule
 QA only    &    24.71       & 38.70  & {\bf 50.00}  \\ 
Subtitle    &   25.16       & 36.45  & {\bf 47.62}  \\
Video        &   27.87      & 36.45  & {\bf 50.67} \\
Videos + subtitle & 25.39    & 40.06  & {\bf 48.87} \\
\bottomrule
\end{tabular}
\caption{Validation experiments with different input modalities and for different word embeddings on best model on MovieQA leaderboard with publicly released code, Layered Memory Network~\cite{Wang2018}. Using subtitles or videos does not improve accuracy. In general, performance differences due to input modalities are dwarfed by the benefits of a better word embedding.
}
\label{table:LMN_results}
\end{table}

\begin{table}[t]
\centering
\small
\setlength{\tabcolsep}{1pt}
\resizebox{\columnwidth}{!}{\begin{tabular}{l@{\hskip 2mm}lccccc} 
\toprule
\# & W2V    &  Movie plots     & Train accuracy  & Train accuracy  & Val accuracy & Val accuracy \\
            & & for training w2v    & (w/o fine-tune)  &  & (w/o fine-tune) &  \\
 
\midrule
1 & \cite{MovieQA}   &    Gen + train + val                     & 27.70  & 41.67  & 26.74  & 38.71\\
2 & \cite{DBLP:journals/corr/MikolovSCCD13}   &    Google News  & 17.84  & 30.40  & 14.56  & 20.31\\
\midrule
 3 & Ours           & Val                    & 20.30 & 24.43 &  40.51 &  41.98\\ 
4 & Ours           & Train                   & 40.19  & 57.46  & 18.39  & 19.30\\
5 & Ours          & Train+val             & 39.90 & 51.64 &   38.48 & {\bf 49.88}\\
6 & Ours           & Gen                & 21.34 & 21.44 &  17.17 & 18.17\\
7 & Ours           & Gen+val           & 21.31 & 27.26 & 34.76 & 36.11\\
8 & Ours           & Gen+train         & 36.77 & 55.33 & 16.59 & 19.63\\
9 & Ours           &  Gen+train+val  & 36.01 & 54.40 & 32.73 & 41.53\\
\bottomrule
\end{tabular}}
\caption{Experiments with our QA only model (for movies+subtitle task) with different amount of movie plots used for training Word2Vec (W2V). This table shows the importance of different word embeddings. Generic word embeddings, like Google's (row 2) gives really poor accuracy. And using a better word embedding (row 5) can give really high accuracy, even without training the QA only model. When we use only val movie plots (row 3) we get good val accuracy but bad train accuracy and vice-versa. Highest accuracy is achieved when we use plots from train+val movies (row 5). Adding movies not in the dataset (row 9), results in degradation of accuracy. Even though same data are used for first and last row, the results differ because of slightly different hyper-parameters.
}
\label{table:2}
\end{table}

\subsection{Ablating the word embeddings}

\begin{figure} \begin{center}
\includegraphics[width=0.49\linewidth,keepaspectratio]{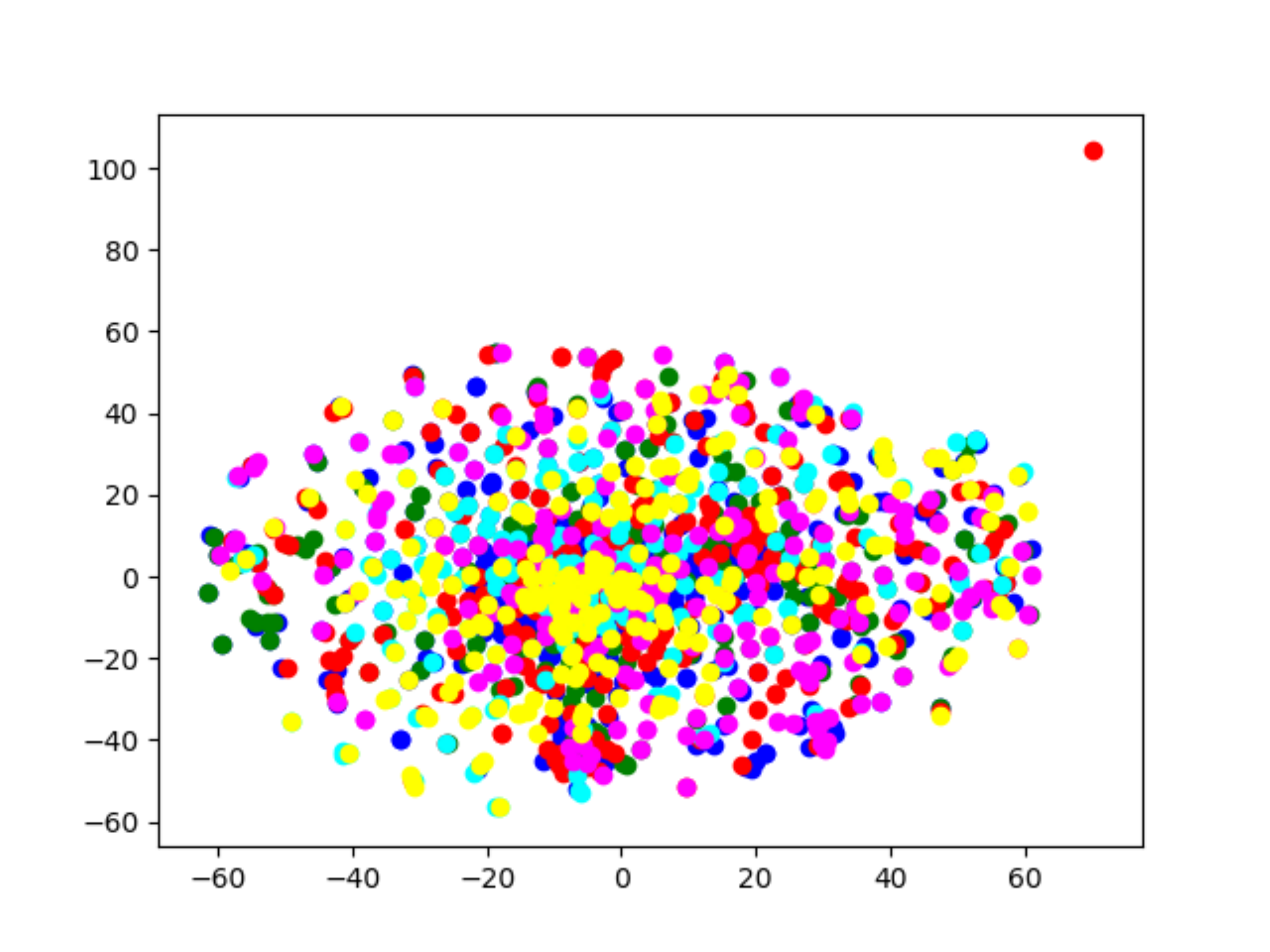}\hfill
    \includegraphics[width=0.49\linewidth,keepaspectratio]{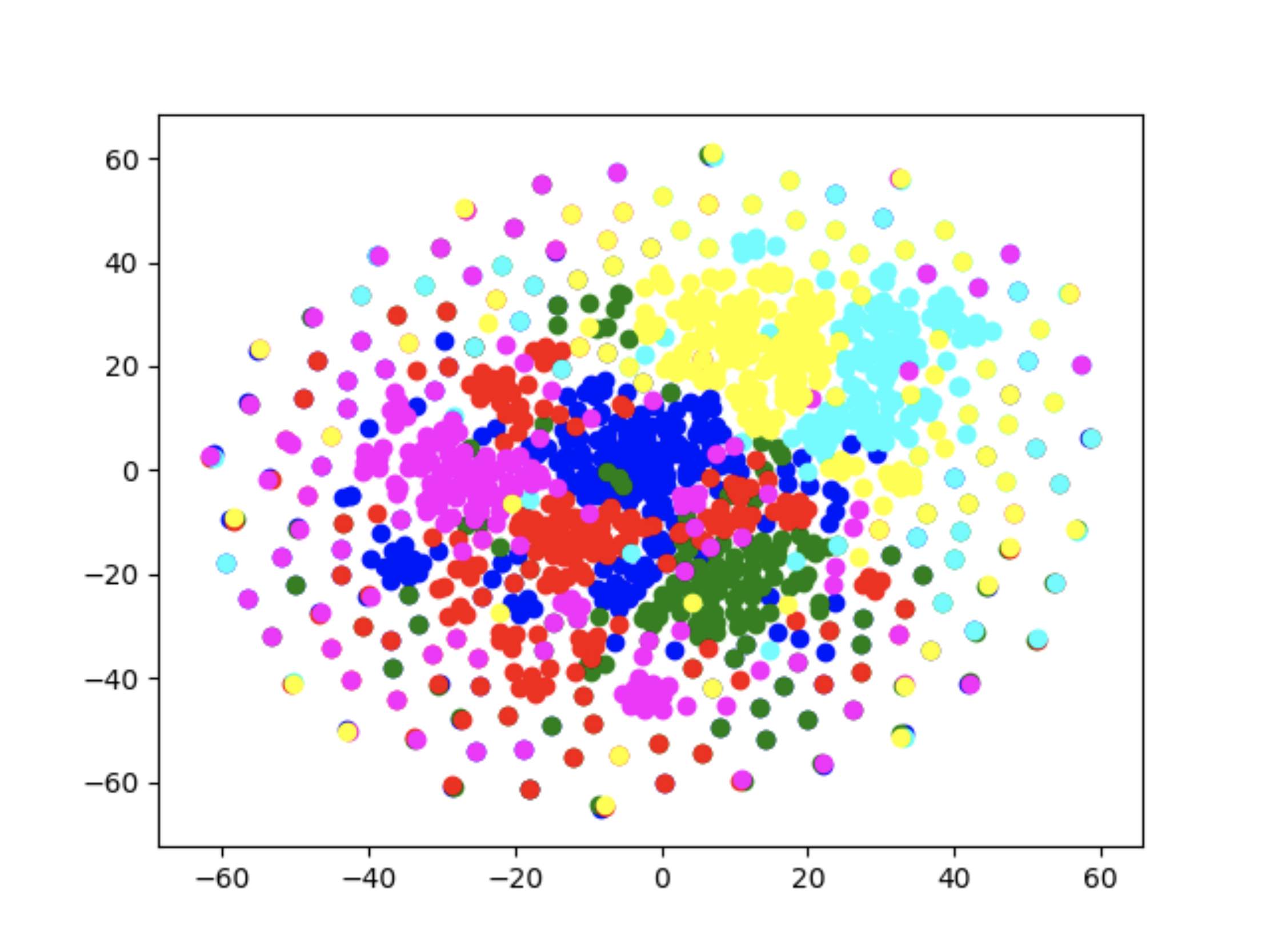}

    \caption{ Left: t-SNE visualization of word embeddings based on google w2v; right: for our WikiWord Embedding w2v. We show them for words taken from 6 different movies, words from the same movies have the same color. For the generic word embedding like google w2v, words from different movies are all jumbled up together and hence they lose the movie-semantics important for this task. In WikiWords, the words from the same movie are clustered together and away from those from other movies. 
}
    \label{fig:tsne_google}
    \end{center}
\end{figure}

{\noindent \bf Movie specific words:} 
We experimented with word2vec (w2v) trained on different data - 1) Google w2v (trained on 100 billion words from Google News dataset, has a vocabulary of 3 million words) 2) MovieQA w2v (provided by the authors, which is trained on about 1400 movie plot synopses including all 408 movies in the MovieQA dataset) 3) Our WikiWords, which is trained on train+test MovieQA plots. Figure~\ref{fig:tsne_google} visualizes Google w2v and WikiWords.
Google w2v is generic and may not contain the names of characters and entities in specific movies. On the other hand, WikiWords tends to embeds words from the same movie together - e.g.\ `Quidditch' and `Harry' refer to the movie Harry Potter. Hence WikiWords captures movie-specific semantics, which is very helpful in answering questions.

{\noindent \bf Google and MovieQA word2vec:} Table \ref{table:2} shows the performance of different w2v's with our QA-only model, evaluated on the train and val set (since submission to the online test server are limited). We plot performance both with and without fine-tuning our linear weighting layer. Google w2v performs poorly and close to chance (20\%, second row), even after fine-tuning.
This is likely because movie-specific words are missing in its vocabulary. MovieQA w2v itself gives about 38.71\% accuracy (first row) after fine-tuning.

{\noindent \bf Our word2vec:} We now explore the effect of using subsets of movie plots to train w2v: 'train', 'val', and 'gen' refer to plots from the train, val, and 1400-(train+val) movies respectively.
Table~\ref{table:2} shows that when including 'val', our QA-only model is able to get high accuracy (40.51\%) even without fine-tuning. This is notable because this system is not trained on {\em any} question-answer pairs.
Finally, Wikiwords (training w2v on 'train+val') leads to the best performance (49.88\%).
Hence, just using plots which are part of the dataset leads to the best accuracy and adding additional movie plots from the general population degrades performance. 

{\noindent \bf Subtitle based word2vec:} As another baseline, we train a w2v with subtitles instead of movie plots from 'train+val' movies and use it in our QA model. This leads to a low accuracy of 26.41\% indicating that w2v trained on subtitles is not able to capture the semantics to exploit the language bias.

\subsection{TVQA dataset experiments}

It is worth exploring the performance of our WikiWord embedding more generally on other datasets. TVQA \cite{lei2018tvqa} is a recent video QA dataset collected from 6 TV series. In contrast to MovieQA, the Mechanical Turkers actually watched the videos (and also read the aligned subtitles) while generating the QAs for TVQA. Since there are no equivalent to movie plots for the TV series, we perform experiments by training word embedding with the subtitles for TVQA dataset. Results in Table \ref{table:TVQA_exp} indicate that although about 40\% of the QA's can be answered without using any context (this result is also mentioned by the TVQA authors), the nature of data used for training word embedding doesn't seem to affect the performance of the QA only models. This shows that TVQA dataset better controls for the biases that MovieQA has.

 \begin{table}[h]
	\centering
	\small
	\setlength{\tabcolsep}{1pt}
	\resizebox{\columnwidth}{!}{\begin{tabular}{ lcc } 
		\toprule
		Model                  & Word embedding  &   Val accuracy \\
		\midrule
		WikiWord embedding     &  Google  News\cite{DBLP:journals/corr/MikolovSCCD13}  & 32.76         \\ 
                                              &  TVQA subtitles & 32.66         \\ 
		\midrule
		TVQA baseline \cite{lei2018tvqa}          & Random weights   & 39.61         \\ 
	                                                                & Wikipedia GLOVE \cite{glove}   & 40.18         \\ 
		                                                            &  TVQA subtitles   & 39.65         \\ 
		\bottomrule
	\end{tabular}}
	\caption{Performance of two QA only models on TVQA dataset - 1) WikiWord embedding model 2) TVQA baseline model \cite{lei2018tvqa} proposed in the paper. For both the models we experiment with word embeddings trained from different data and observe that the performance doesn't change.}
	\label{table:TVQA_exp}
\end{table}

\section{Conclusion}

We show that the MovieQA dataset has language bias and present a simple QA only model that exploits it. 
Our key idea is to train the word2vec model on a {\em subset} of the data used by state of the art methods, by focusing only on the train and test movie plots.
This model achieves
state of the art performance on four of the five categories on the leaderboard at the time of submission.

{\scriptsize
\bibliographystyle{ieee}
\bibliography{refs}
}

\clearpage

\appendix

\section{Appendix}

\subsection{What does WikiWords embeddings learn?}
\begin{figure*} [h!]
    \begin{center}
\includegraphics[width=\linewidth,keepaspectratio]{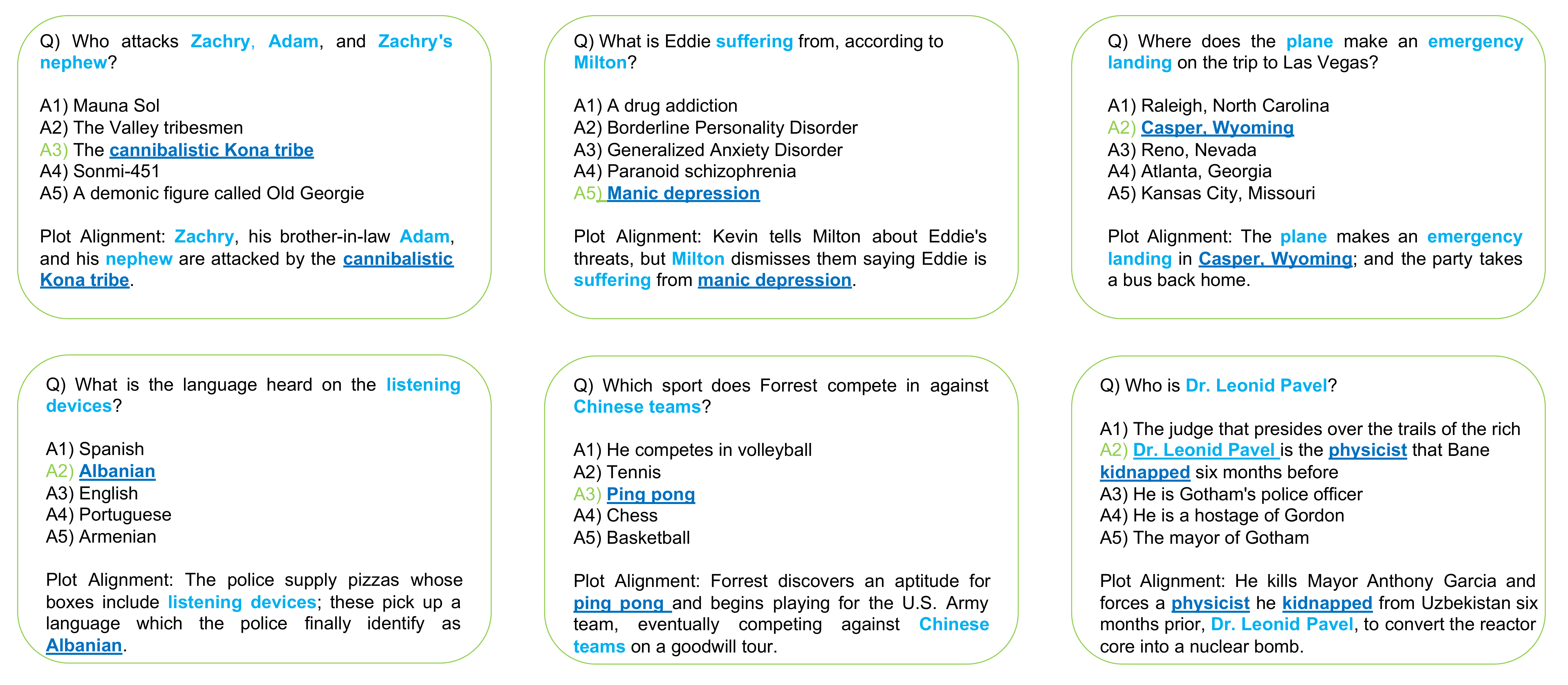}
\end{center}
    \caption{{\bf QAs which are correctly predicted by our simple QA only model and hence are the biased QAs.} Correct answer is highlighted in green. Light blue coloured words are the movie specific words common between the question and the line in movie plot from which the question was made by Amazon Mechanical Turkers. Dark blue underlined words are the movie specific words common between the correct answer and the line in the movie plot. For example for the question in 2nd column, the model predicted A3) because `Ping pong' (movie specific word in the correct answer) is the only word that appears close to `Chinese teams' (movie specific word in the question) in the movie plot. For the incorrect answers options, the words in them don't occur in the movie plot and hence these options are very different semantically in word embeddings. Due to this reason, it's easy to find the correct answer without using any other information like from videos, hence making it a biased QA.}
    \label{fig:biased}
\end{figure*}

\begin{figure*} [h!]
    \begin{center}
\includegraphics[width=\linewidth,keepaspectratio]{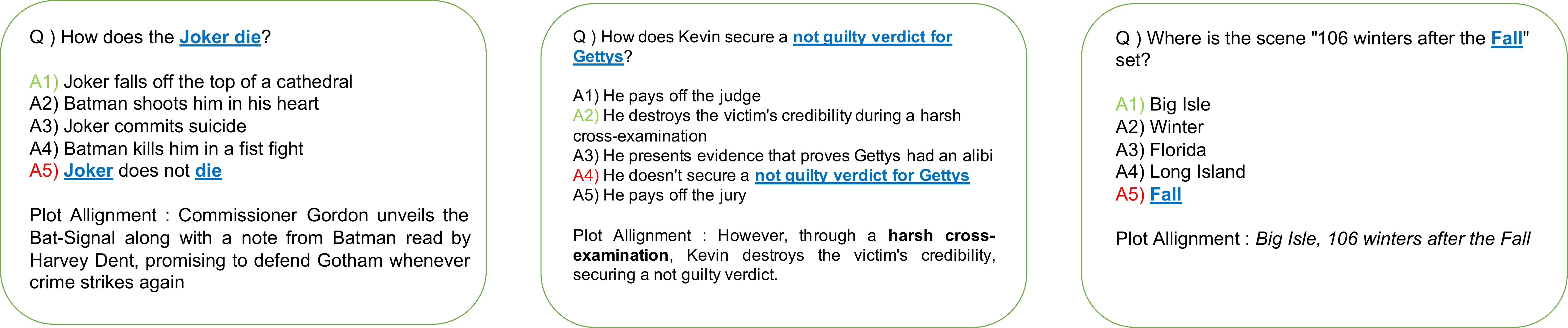}
    \end{center}
    \caption{{\bf QAs which are wrongly predicted by our simple QA only model.} These are the QAs which are more likely to be less biased. Prediction of the model is in red and the correct answer is in green. For example in the third question, the model predicted A5. This is because amongst all the answer choices the word `Fall' in A5) is the only common movie specific word amongst the words in the question. Hence A5) would have very high dot product similarity in the word embedding space with the question and so the model predicted it as the answer.}
    \label{fig:unbiased}
\end{figure*}

In a way our simple QA model with the pre-trained word2vec model is trying to memorize the occurrence of nearby words in the movie plot synopsis. Since the question-answers are made by AMT workers by only looking at the movie plot synopsis, it is able to correctly answer the QAs in half the dataset.  Figure~\ref{fig:biased} shows the predictions of our simple model with `train+val' word2vec which are correct and Figure~\ref{fig:unbiased} shows the predictions which are incorrect. It also highlights the prominent words in the question, the correct answer and the line in the movie plot form which the QA was made by the AMT workers.

We find that the model first tries to select the answer choice which has the highest number of movie specific words as that in the question. This happens because in this case, the word embedding of question and the selected answer would be very close. Another aspect of our model is to select the answer whose movie specific word(s) occur adjacent to the movie specific word(s) of the question in the movie plots (since in word2vec space, nearby text words have very high dot product similarity). Again this ensures that the word embeddings of question and the selected answer choice would have very high similarity. And surprisingly just doing this, our simple model achieves close to 50 percent accuracy on the video-based QA task with only looking at question and picking the answer.

\subsection{Towards an Unbiased Dataset: Easy-question Removal}

 \begin{table}[h]
	\centering
\begin{tabular}{ lcc } 
		\toprule
		Type                  & Our model  & TVQA baseline \cite{lei2018tvqa}   \\
		\midrule
		Original dataset     & 49.88    & 32.50         \\ 
Only biased          & 99.41    & 47.80         \\ 
Only unbiased        & 25.68    & 22.50         \\ 
\bottomrule
	\end{tabular}
	\caption{Comparison of performance on different splits of MovieQA dataset for 2 different QA only models. The first row shows the original dataset. The second row shows the subset of original dataset which is biased i.e. our QA only model is able to correctly answer them. The third row is the subset which our QA only model is unable to correctly answer, resulting in chance level accuracy. These are the unbiased QAs and hence is the hardest split, and which would need information from videos and subtitles. 
}
	\label{table:reduce_bias}
\end{table}

Our method naturally generates a partition of the data that is free of trivial language biases and can potentially be used for further progress in video-language modeling. We consider the QA's which can be correctly answered by our WikiWord  embedding model as the biased QA's and the rest as the unbiased QA's. In order to ensure that we don't overfit to the data when finding these biased questions from the training set, we use the predictions of our {\em untrained} model. That is we take our QA only model which uses the pretrained word2vec, and we do not further train it with question-ground truth answer pairs. As per the Table \ref{table:2} we achieve around 40\% accuracy on train and val sets and so we drop all these QA's. Our hypothesis is that these questions are the really easy and the most biased ones, and video based models can solve without actually needing context from videos. Table \ref{table:reduce_bias} compares performance when QA only models are trained and tested with the original dataset, only the biased subset, only the unbiased subset. To ensure that our unbiased dataset is competitive for multiple models, we show the performance of our QA-only model and that of the baseline QA-only model proposed in TVQA \cite{lei2018tvqa} dataset. We observe the unbiased subset is harder for both models and results in close to chance level accuracy (20\%). The unbiased subset hence provides a quick fix to the dataset.

\subsection{Performance of our model for different amount of movie plots used for training word2vec}

\begin{figure} [h!]
	\begin{center}
		 \includegraphics[width=\linewidth,keepaspectratio]{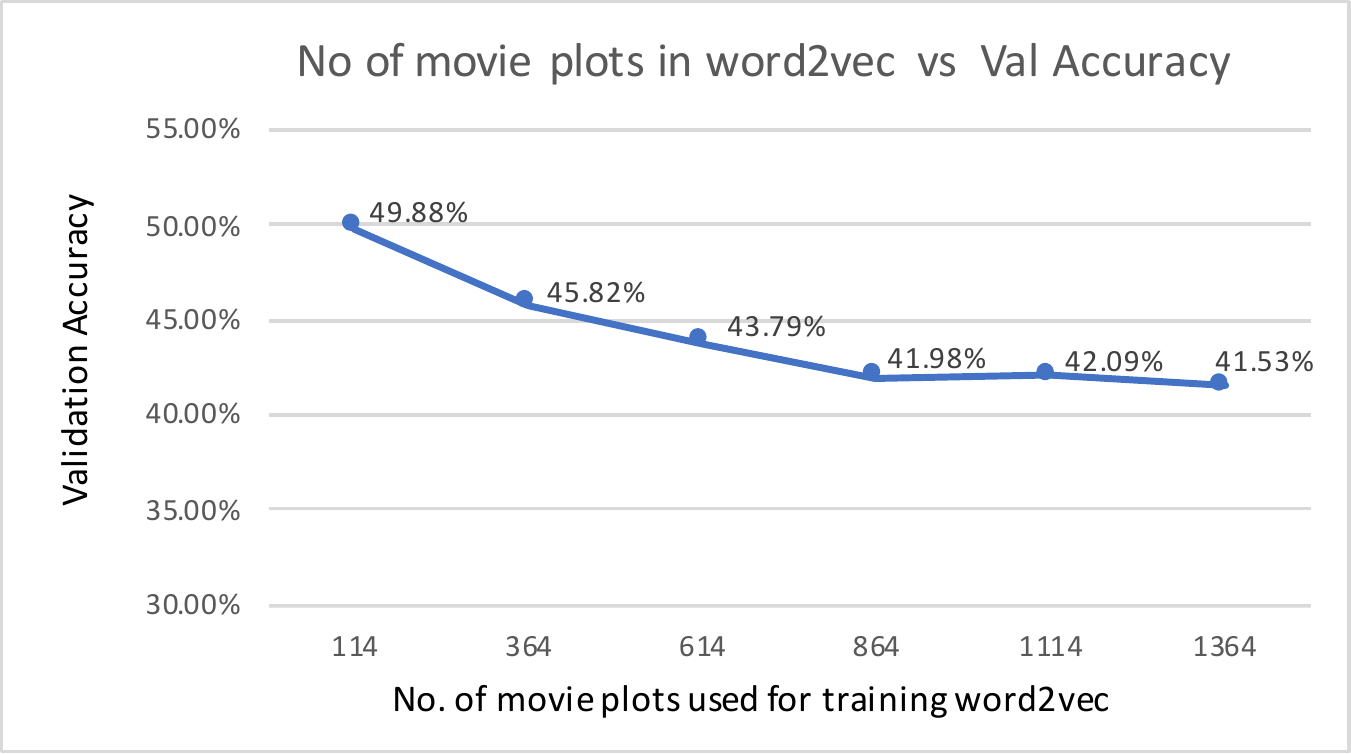}
	\end{center}
	\caption{{\bf Adding additional plots reduces overall performance.} We now explicitly evaluate the accuracy of our model on the validation set for different amount of movie plots used for training word2vec. The left most data point (114 plots) corresponds to word2vec trained with movie plots just from train+val set which is used in our WikiWord embedding model. On adding additional movie plots from random movies till we reach the right most data point which corresponds to word2vec provided by MovieQA authors, which is trained on a total of 1364 movie plots. The plot indicates adding movie plots for training word2vec degrades the accuracy of our model.}
	\label{fig:movei_plot_accuracy}
\end{figure}

\end{document}